\documentclass[twocolumn]{svjour3}          
\smartqed  
\usepackage[usenames]{color}
\usepackage[numbers]{natbib}
\usepackage[table]{xcolor}
\usepackage{amssymb}
\usepackage{makecell}
\usepackage{multirow}
\usepackage{rotating}
\usepackage{array}
\usepackage{times}
\usepackage{graphicx}
\usepackage{subfigure}
\usepackage{enumitem}
\usepackage{rotating}
\usepackage{xurl}
\usepackage{rotating, graphicx}
\usepackage{amsmath}
\usepackage{booktabs}
\usepackage{lscape}
\usepackage{verbatim}
\usepackage{geometry}
\usepackage[misc]{ifsym}
\usepackage{algorithm}
\usepackage{algpseudocode}
\papertype{generic article}
\usepackage[pagebackref=true,breaklinks=true,linkcolor=red,anchorcolor=green, citecolor=blue,letterpaper=true,colorlinks,bookmarks=false]{hyperref}
\usepackage{breakurl}
\usepackage{graphicx}
\definecolor{DarkBlue}{rgb}{0,0,1}
\definecolor{DarkRed}{rgb}{0.6,0.00,0.08}
\definecolor{DarkGreen}{rgb}{0.0,0.6,0.08}
\definecolor{LightBlue}{rgb}{0.88,0.92,0.95}
\definecolor{Orange}{rgb}{1,0.75,0}
\usepackage{xspace}
\makeatletter
\DeclareRobustCommand\onedot{\futurelet\@let@token\@onedot}
\def\@onedot{\ifx\@let@token.\else.\null\fi\xspace}
\def\eg{\emph{e.g}\onedot} 
\def\ie{\emph{i.e}\onedot} 
 
\def\etc{\emph{etc}\onedot}

\makeatother

\newcommand{\myparagraph}[1]{\vspace{6pt}\noindent{\bf #1}}

\begin{document}

	\title{Imagine the Unseen: Occluded Pedestrian Detection via Adversarial Feature Completion}
	\subtitle{}

	\author{Shanshan Zhang \and Mingqian Ji \and Yang Li \and Jian Yang*  
	}
	
	\authorrunning{Shanshan Zhang \emph{et al.}} 

	\institute{\{shanshan.zhang,csjyang\}@njust.edu.cn \\
		Nanjing University of Science and Technology, China \\     
	}
	\date{Received: date / Accepted: date}

	\maketitle
	
	\begin{abstract}
		Pedestrian detection has significantly progressed in recent years, thanks to the development of DNNs. However, detection performance at occluded scenes is still far from satisfactory, as occlusion increases the intra-class variance of pedestrians, hindering the model from finding an accurate classification boundary between pedestrians and background clutters. From the perspective of reducing intra-class variance, we propose to complete features for occluded regions so as to align the features of pedestrians across different occlusion patterns.
		An important premise for feature completion is to locate occluded regions. From our analysis, channel features of different pedestrian proposals only show high correlation values at visible parts and thus feature correlations can be used to model occlusion patterns.	
		In order to narrow down the gap between completed features and real fully visible ones, we propose an adversarial learning method, which completes occluded features with a generator such that they can hardly be distinguished by the discriminator from real fully visible features.	
		We report experimental results on the CityPersons, Caltech and CrowdHuman datasets. On CityPersons, we show significant improvements over five different baseline detectors, especially on the heavy occlusion subset. Furthermore, we show that our proposed method FeatComp++ achieves state-of-the-art results on all the above three datasets without relying on extra cues.
		
		\textbf{Index Terms}\textemdash
		Pedestrian detection; occlusion handling; feature completion; adversarial learning.
		
	\end{abstract}

	\section{Introduction}
	Pedestrian detection is an important problem in computer vision, serving as a basic building block in many intelligent systems such as autonomous driving, video surveillance and robotics. During the past few years, the detection performance has been improved largely thanks to the development of DNNs \cite{Hosang2015Cvpr,shanshan-pami2017}. Although the progress is significant across different datasets, we notice the performance is still far from satisfactory when pedestrians are occluded, which happens rather frequently in practice, \eg a pedestrian may be occluded by another pedestrian walking side by side in urban traffic scenes, or by some pole standing on the street.
	Therefore, it has been agreed by the community that occlusion handling is an urgent problem to solve for pedestrian detection.
	Recently, some new datasets \cite{CityPersons,shao2018crowdhuman} consisting of rich occlusion scenes have been proposed to encourage more works investigating how to improve the performance under occlusions.
	
\begin{figure}
		\centering
		\includegraphics[width=0.4\textwidth] {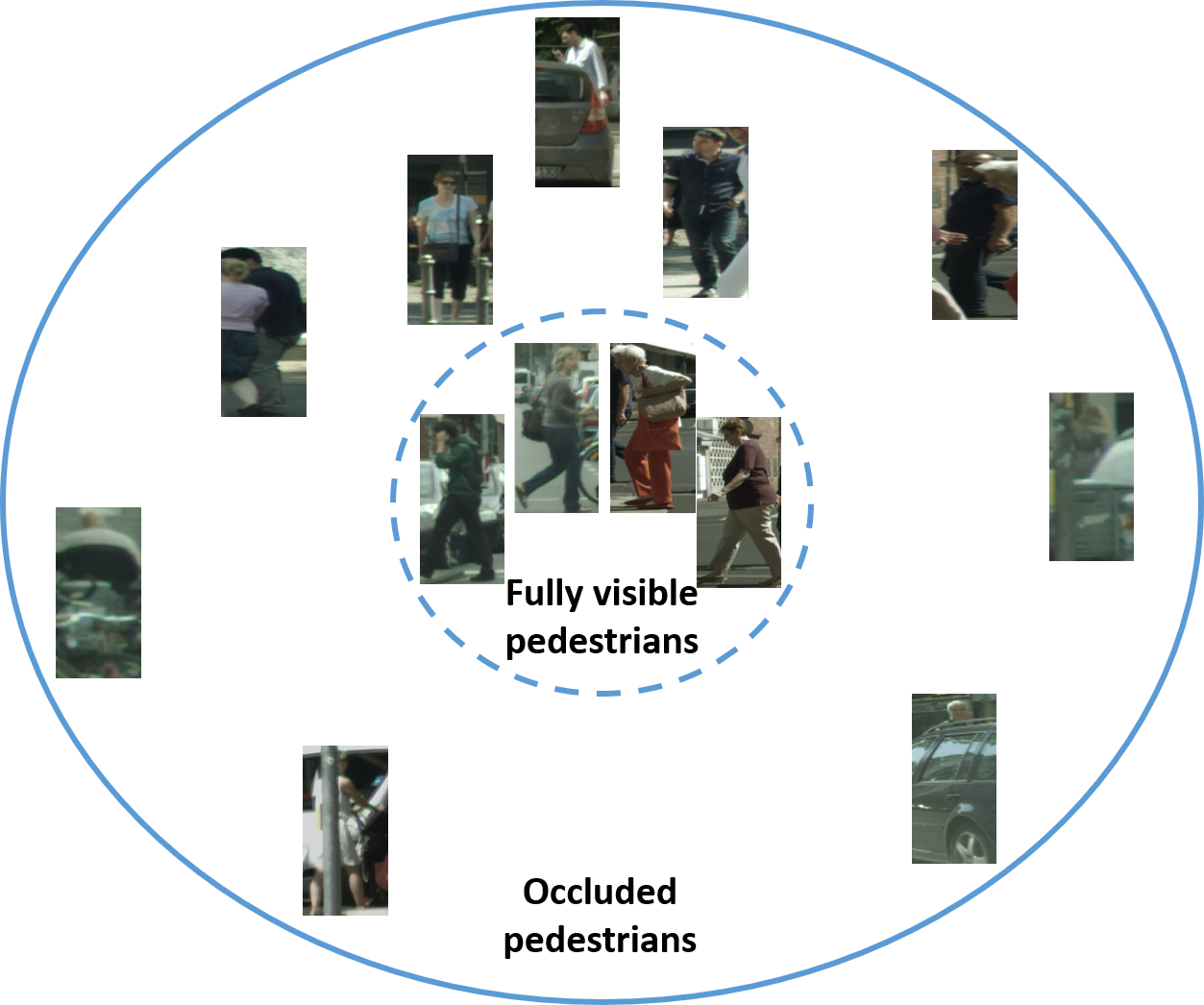}
		\caption{Various occlusion patterns result in large intra-class variance of pedestrians.}
		\label{fig:feature_space_toy}
	\end{figure}

	It is worth thinking why occlusion makes detection more difficult before delving into the solutions. We assume the reason is that occlusion increases the intra-class variance of pedestrians w.r.t. appearance, \ie the difference between fully visible and occluded pedestrians is obviously huge caused by the occluding objects or pedestrians, as shown in Fig. \ref{fig:feature_space_toy}.
	This intra-class variance has a negative impact on the detection accuracy as the sparse feature space of pedestrians would hinter the model from finding an accurate classification boundary between pedestrians and background clutters.
	Thus, it is not surprising to observe for most pedestrian detectors the detection performance drops as occlusion ratio increases.
	In this sense, it would be an effective way of solving occluded pedestrian detection to reduce the pedestrian intra-class variance caused by occlusion, \ie to align the features of pedestrians, regardless of occlusion level.

	A previous attempt from this perspective can be found in \cite{zhou2019feature-transformation-pedestrian}, where a simple feature transformation is applied to close the gap between occluded and non-occluded pedestrians by forcing occluded pedestrian samples approach the centroid of easily classified non-occluded ones.
	However, the feature transformation is implicit and blind to occlusion patterns, and thus the improvement is rather limited.
	In contrast, in this paper we propose a more explicit way called feature completion. We first locate the occluded regions and then imagine the unseen parts by borrowing features from the corresponding regions of fully visible pedestrians. After such completion, the features of occluded pedestrians are expected to align with those of fully visible ones, making the entire feature space much more compact. Our method is rather intuitive but there are two key problems to solve: 
	
	\textbf{(1) How to locate occluded regions?}
	To locate occluded regions, we need to model various occlusion patterns in an effective way. Most approaches pre-define occlusion patterns with a combination of a rectangular visible box and a full body box, and then learn corresponding representations for each pattern  \cite{DeepParts,Enzweiler2010Cvpr,Ouyang2012Cvpr,Mathias2013Iccv,MultiLabel-occl,Ouyang2013ICCV-joint}. More recently, it has been proposed to use part detections as fine-grained guidance to allow the representations to focus more on visible parts \cite{AttentionNet} in a dynamic way. However, from our analysis of occluded scenes, we find the visible boxes cannot accurately model complex occlusion models while part detections fail to model person-person occlusions.
	In order to better handle various occlusion patterns, we propose an alternative way by exploiting correlations between pedestrians, and find the convolutional features of them correlate well for visible parts. This suggests feature correlations can be used as an effective way of modeling occlusions, without relying on extra cues, \eg visible boxes or part detections.
	
	\textbf{(2) How to close the gap between artificially completed features and real fully visible ones?}
	After locating occluded regions, we perform feature completion with the help of fully visible pedestrians.
	However, it would induce severe artifacts if we simply borrow features from fully visible pedestrians to replace the original features at occluded regions.
	These artifacts would result in a high level of intra-class variance, which is harmful for the following classification procedure.
	Aiming for a compact feature space, we need to push completed features towards fully visible ones. To this end, we propose an adversarial learning method, where the generator for feature completion is optimized such that the completed features can hardly be distinguished by the discriminator from real fully visible features.
	
	
	\subsection{Contributions}
	In summary, our contributions are as follows:
	
	(1) We provide an analysis on occlusion pattern modeling, and find feature correlations between pedestrians are particularly effective for various occlusion patterns without relying on extra cues or annotations, \eg visible boxes or part detections.
	
	(2) Given the located occluded regions, we propose a new progressive adversarial feature completion method. At each step, features of occluded samples are fed into a generator, where we borrow features from fully visible prototypes for occluded regions and refine the entire feature maps, to align with real fully visible ones.
	Synthetic and real occluded samples are used to construct adversarial pairs at two steps, respectively, allowing for more effective step by step optimization.

	(3) Our proposed method achieves consistent improvements over five different baselines. In particular, on the CityPersons heavy occlusion subset, the improvement is up to $\sim$7pp.
	Moreover, we achieve state-of-the-art results on three different datasets, \ie  CityPersons, Caltech and CrowdHuman.
	
	\subsection{Related Work}
	Since we use DNN detectors as our baselines, and employ feature correlations to model occlusion patterns, we review recent work on DNN based pedestrian detectors, occlusion handling for pedestrian detection and object correlations respectively.
	
	\myparagraph{Pedestrian detection with DNNs.}
	Deep neural networks (DNNs) have achieved great success on the pedestrian detection task. 
	Early works \cite{Hosang2015Cvpr,shanshan_cvpr16} are based on the RCNN structure \cite{rcnn14CVPR}, which relies on high-quality external proposals to achieve good performance. 
	Instead, the FasterRCNN detector \cite{renNIPS15fasterrcnn} enables end-to-end training without relying on external proposal methods. Thus, Zhang et al. \cite{CityPersons} adapts the standard FasterRCNN detector to the pedestrian detection task via making some modifications including setting fine-grained anchor scales, finer feature stride, \etc, to improve the performance significantly. Also, it has been highlighted the diversity of training data is beneficial to enhance models' generalization ability \cite{CityPersons}. Furthermore, Hasan et al. \cite{Ped_Det_Elephant} show that proper training strategies are important when training with a mixture of datasets.
	Some other works focus on small scale pedestrian detection \cite{Graininess,SA-FastRCNN,TopologicalLine}. 
	For example, SA FastRCNN \cite{SA-FastRCNN} introduces multiple built-in sub-networks, each of which copes with pedestrians from a specific scale range, and then adaptively combines all outputs to generate the final detection results; TLL \cite{TopologicalLine} integrates somatic topological line localization and temporal feature aggregation for detecting multi-scale pedestrians.
	On the other hand, some attempts have been made to employ the single-stage detection architecture for pedestrian detection \cite{ALF} aiming for high inference speed, but the performance is inferior to those detectors \cite{cai16mscnn,CityPersons,brazil2019autogressive-pedestrian} based on the two-stage FasterRCNN architecture \cite{renNIPS15fasterrcnn}.
	To avoid manual design of anchors, anchor-free detectors have become popular, but they usually need to be fed with more training data and require a longer time to converge \cite{liu2019anchor-free-pedestrian}.
	More recently, Transformer based detectors including DETR \cite{DETR} and its variants show promising results, getting rid of the post-process of Non Maximum Suppression (NMS). 
	In this paper, for most experiments, we follow \cite{CityPersons} and use the adapted FasterRCNN architecture as the baseline detector, which is easy to re-implement and achieves comparable results to state-of-the-art detectors; we also choose Deformable DETR \cite{D-DETR} as a different kind of baseline detector to verify the generalization ability of our method.
	Need to mention that, in principle our proposed method can be applied on top of any arbitrary DNN based detector.

	\myparagraph{Occlusion handling for pedestrian detection.} 
	Most previous methods make use of additional visible bounding boxes to infer occlusion patterns. These works include ensemble models \cite{Enzweiler2010Cvpr,Mathias2013Iccv,Ouyang2012Cvpr,DeepParts}, and joint inference of various occlusion patterns \cite{MultiLabel-occl,Ouyang2013ICCV-joint,Bi-box}. Some recent works also find that it is beneficial to allow the feature representations to focus more on the visible regions by using visible boxes or part detections as additional guidance \cite{AttentionNet,OcclusionAware,Bi-box}.
	Other works show that it is helpful to use a weak segmentation loss as additional supervision so as to improve performance \cite{SDS-RCNN,NohCVPR18,pang2019mask-attention-pedestrian}. 
	It is also shown that a proper feature transformation can be applied to close the gap between occluded and non-occluded pedestrians \cite{zhou2019feature-transformation-pedestrian}.
	Li et al. \cite{HGPD} enhance the features for occluded pedestrians by learning the inter-proposal and intra-proposal affinities using a hierarchical graph.
	The above methods usually deal with general occlusion, while some other works focus on occlusion brought by nearby persons, targeting the crowded scenes.
	For example, RepLoss \cite{RepulsionLoss} and OR-CNN \cite{OcclusionAware} design new loss functions to better separate proposals belonging to different persons close to each other.
	Besides, some works show that a proper Non Maximum Suppression (NMS) is important to boost performance at crowded scenes. In \cite{liu2019adaptive-nms-pedestrian}, an adaptive IoU threshold is chosen based on density prediction; and in \cite{PBM}, the NMS is performed on paired visible boxes instead of full body boxes.
	Other works go beyond single image by exploiting additional information to help occluded pedestrian detection, \eg temporal \cite{TCE, TopologicalLine}, depth \cite{W3Net} or vision-language pretraining models \cite{VLPD_2023_CVPR}, yet we believe it is still worth investigating to push the limits of reasoning based on single images.
	In this work, we deal with general occlusion and explore how to model various occlusion patterns without relying on additional cues or annotations regarding visible parts. Our proposed method further generates reasonable features for occluded pedestrians so as to close the gap between fully visible and occluded pedestrians.
	
	\myparagraph{Object correlations in DNNs.}
	Object correlations play an important role in the area of object detection.
	Early on, DPM \cite{DPM} uses co-occurrence to indicate how likely two object classes can exist in one image.
	However, in the era of DNNs, there are very few works investigating object correlations since the receptive fields are large and thus incorporate contextual information. Recently, some efforts have been made to model object correlations in different ways.
	In \cite{Chen2017CVPR}, a new sequential reasoning architecture is proposed to exploit object-object relationships to sequentially detect objects in an image;
	in \cite{SIN}, a structure inference network (SIN) is proposed to reason object state in a graph, where a memory cell is used to encode messages from scene
	and other objects into object state; 
	it has also been shown that properly designed global relation features enhance the appearance features
	\cite{RelationNetworks}.
	In this work, we exploit the correlations between an occluded detection proposal and a fully visible one to model various occlusion patterns.

	\begin{figure*}[h]
		\centering
		\includegraphics[width=\textwidth] {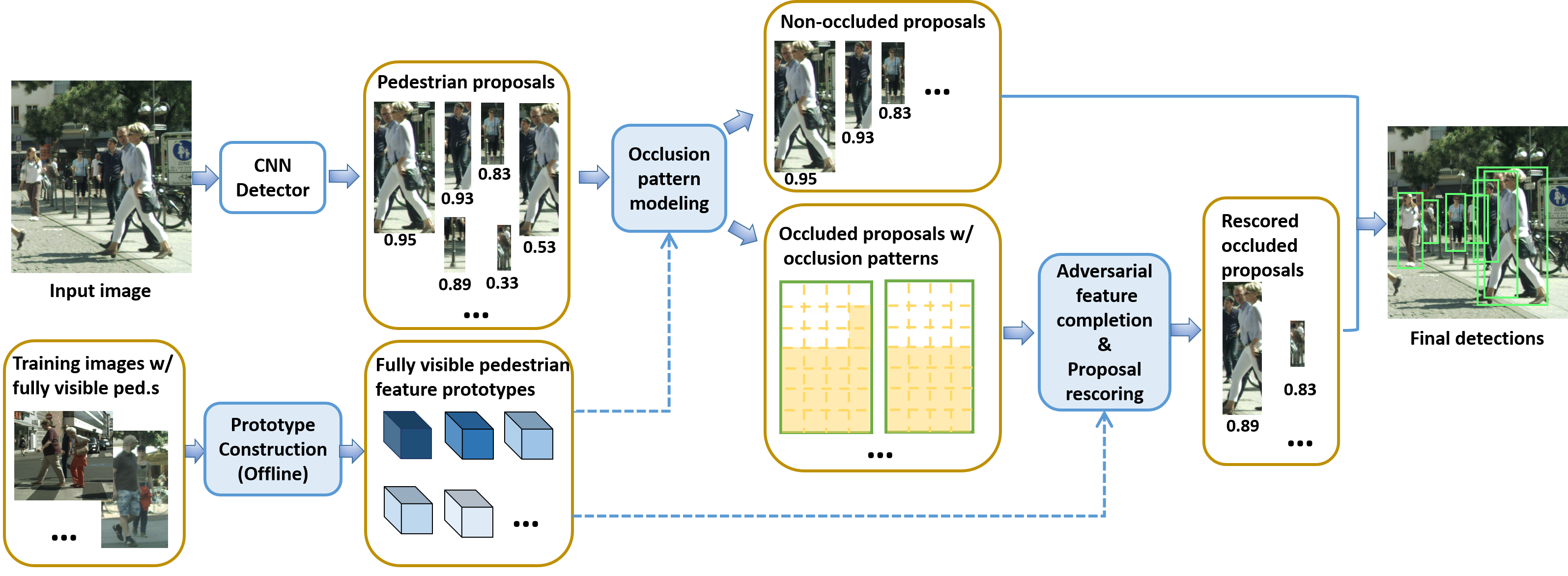}
		\caption{Overview of our proposed approach. Offline procedure: to construct feature prototypes for all fully visible pedestrians in the training set. Online procedure: a baseline detector (\eg Faster RCNN), to generate pedestrian proposals; an occlusion pattern modeling module, to distinguish occluded proposals from non-occluded ones and also to provide occlusion patterns for those occluded ones; an adversarial feature completion module, to generate features for occluded proposals via an adversarial learning framework; finally, each occluded proposal is rescored based on the completed features and the final detections include both non-occluded proposals from the baseline detector and rescored occluded ones.}
		\label{fig:overview}
	\end{figure*}
	
	\section{Approach}\label{sec:approach}
	In this section, we first give an overview of our proposed approach, followed by detailed descriptions to its three key components: fully visible pedestrian feature prototype construction, occlusion pattern modeling, and adversarial feature completion at occluded regions.

	\subsection{Overview}
	An overview of our proposed approach is shown in Fig. \ref{fig:overview}. First, we have an offline procedure to construct feature prototypes based on all fully visible pedestrians in the training set, which will be used in the other two key modules: (1) to compute correlations so as to locate occluded regions in the occlusion pattern modeling module; (2) to serve as base features for completion in the adversarial feature completion module.
	
	For online training/testing, each image is first fed into a baseline detector (\eg FasterRCNN) to generate pedestrian proposals, which may contain fully visible pedestrians, occluded pedestrians and background clutters;
	then, an occlusion pattern modeling module is applied by computing a correlation map between each proposal and its nearest fully visible pedestrian feature prototype, so as to distinguish occluded proposals from non-occluded ones and also to provide occlusion patterns for those occluded ones;
	after that, each occluded proposal goes through an adversarial feature completion module, where features for occluded regions are borrowed from the nearest fully visible prototype and further refined via a progressive adversarial learning framework, such that the completed features are similar to real fully visible ones;
	finally, each occluded proposal is rescored based on the completed features and the final detections include both non-occluded proposals from the baseline detector and re-scored occluded ones.

	\subsection{Fully Visible Pedestrian Feature Prototype Construction}
	Fully visible pedestrian feature prototypes are constructed offline.
	All fully visible samples in the training set are collected to train a baseline detector; after convergence, we build a feature pool containing channel features after the RoIPooling layer, each with a dimension of $C\times X\times Y$, where $C$ indicates the number of channels and $X, Y$ indicate the numbers of pooling grids horizontally and vertically, respectively.
	We then cluster all features into $K$ groups using standard K-means algorithm, and take each cluster center as a feature prototype.
	
	In order to understand the differences among the above $K$ prototypes, we observe the samples in each cluster and find they differ significantly w.r.t. scale. Take the FasterRCNN detector with $K=4$ for example, the scale mean and std values for $4$ clusters are as follows: $64\pm9.44, 105\pm19.33, 181\pm36.62, 340\pm131.33$.
	This finding offers a rather convenient and efficient way to find the nearest prototype for each proposal, \ie computing the distance between the current proposal's scale and the above mean values. This is much cheaper than computing distances in the high-dimensional feature space.

	\subsection{Occlusion Pattern Modeling}
	\label{sec:Occlusion Pattern Modeling}
	
	\begin{figure}
		\centering
		\subfigure[]{\includegraphics[width=0.325\textwidth] {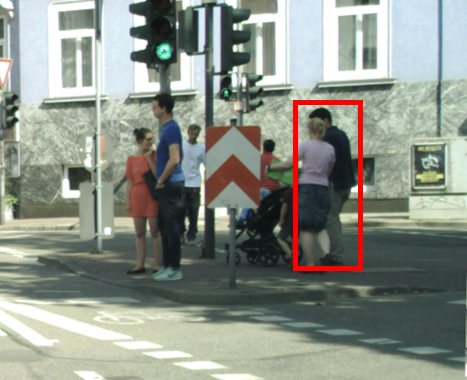}\label{fig:occluded_person}} 
		\hspace{10pt}
		\subfigure[]{\includegraphics[width=0.11\textwidth]{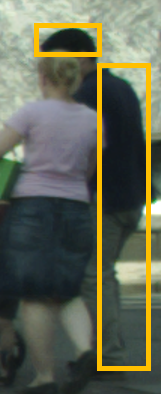}\label{fig:occluded_person_vbb}} 
		\hspace{10pt}
		\subfigure[]{\includegraphics[width=0.5\textwidth]{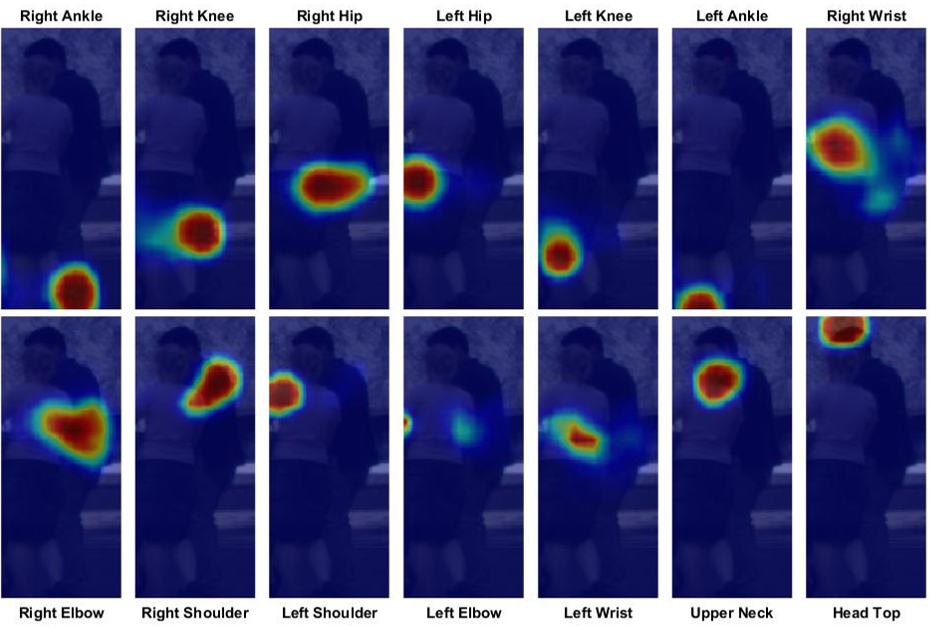}\label{fig:occluded_person_parts}}
		
		\caption{Person-person occlusion pattern shown in (a) can hardly be properly modeled by one rectangular box (b) or body part detections (c).
		}
		\label{fig:occlusion patterns}
	\end{figure}

	On most standard pedestrian datasets, a combination of visible box and full extent box is used to model occlusion patterns. However, one single rectangular box is not able to cover the entire visible area when the occlusion is irregular, especially introduced by a nearby human body. For example, as shown in Fig.~\ref{fig:occluded_person_vbb}, Bob is occluded by Alice next to him; we need at least two rectangular boxes to cover the visible parts of his body. 
	However, such compound visible boxes are expensive to annotate and are hard to precisely cover the visible parts, and thus are not readily available for standard datasets such as Caltech and CityPersons. 
	
	Alternatively, body part detections can be used to model irregular occlusion in a more fine-grained way as proposed in \cite{AttentionNet,AttentionNetIJCV}. This is true for person-object occlusion; however, when the occlusion is caused by another person, the part detections from the occluder tend to mix up with those from the occluded person, thus resulting in confusion.
	In Fig.~\ref{fig:occluded_person_parts}, we show 14 body part heatmaps generated by a body part detector \cite{deepercut}.
	We can see in the box of the occluded person that each body part has high scores, many of which are in fact triggered by the occluder. Thus it is unclear how to effectively employ body part detections for person-person occlusion modeling.

	Given the above discussion, it is not promising to use either visible boxes or part detections to identify diverse occlusion patterns. Therefore, we propose an alternative way of exploiting feature correlations between pedestrian proposals.
	
	\subsubsection{Feature correlations}
	
\begin{figure}
		\centering
		\includegraphics[width=0.4\textwidth] {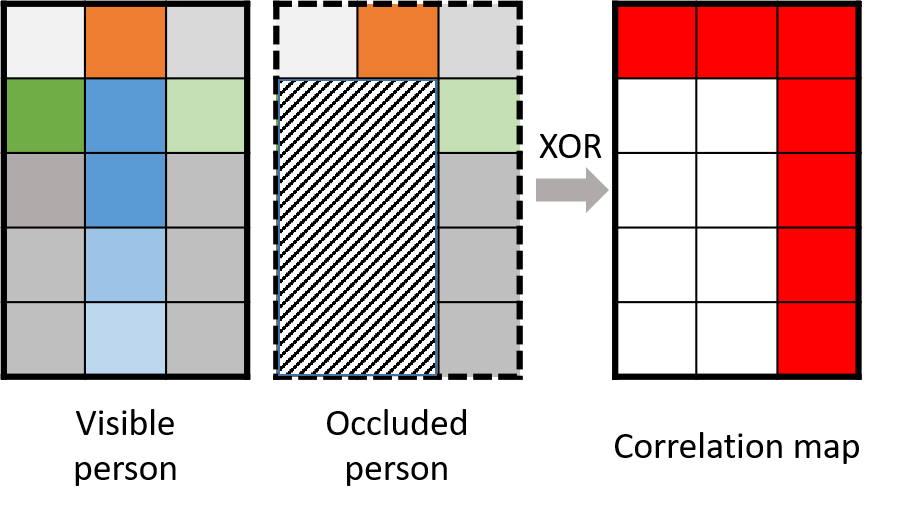}
		\caption{A toy example illustrating how the correlations between visible and occluded pedestrians can be used to model occlusion patterns. Different colors indicate different body parts. Red in correlation map indicates high correlation values shown at visible parts only.}
		
		\label{fig:correlation_toy_example}
	\end{figure}

	We first show the correlations between two pedestrian proposals in Fig. \ref{fig:correlation_toy_example} via a toy example. The occlusion mask on the occluded pedestrian is randomly generated. We assume the features at corresponding parts of different pedestrians are the same (shown with the same color) and then we perform XOR operation to generate a correlation map, where we can see the regions with high responses (shown in red) perfectly represent the visible parts of the occluded pedestrian.
	Therefore, we conclude the correlations between the visible and occluded pedestrians are potential to model occlusion patterns.
	
	Next, we further analyze the correlation of two pedestrian proposals' convolutional channel features.
	First, we pick two proposals that are highly overlapped with Alice (visible) and Bob (occluded) respectively.
	Since many channels from a typical DNN based pedestrian detector represent body parts \cite{AttentionNet}, we pick those interpretable channels for observation, each of which represents one body part.
	Then we crop the channel features based on Alice's and Bob's proposals, and compute the correlation map $c$ between the two feature maps element-wise as follows:
	\begin{equation}
		c(x,y)=\frac{f_{vis}(x,y)  f_{occ}(x,y)}{1+|f_{vis}(x,y) - f_{occ}(x,y)|},
	\end{equation}
	where $f_{vis}$ and $f_{occ}$ are feature matrix for Alice (visible) and Bob (occluded) proposals, respectively; $x,y$ are horizontal and vertical indexes.
	
\begin{figure}
		\centering
		\includegraphics[width=0.5\textwidth]{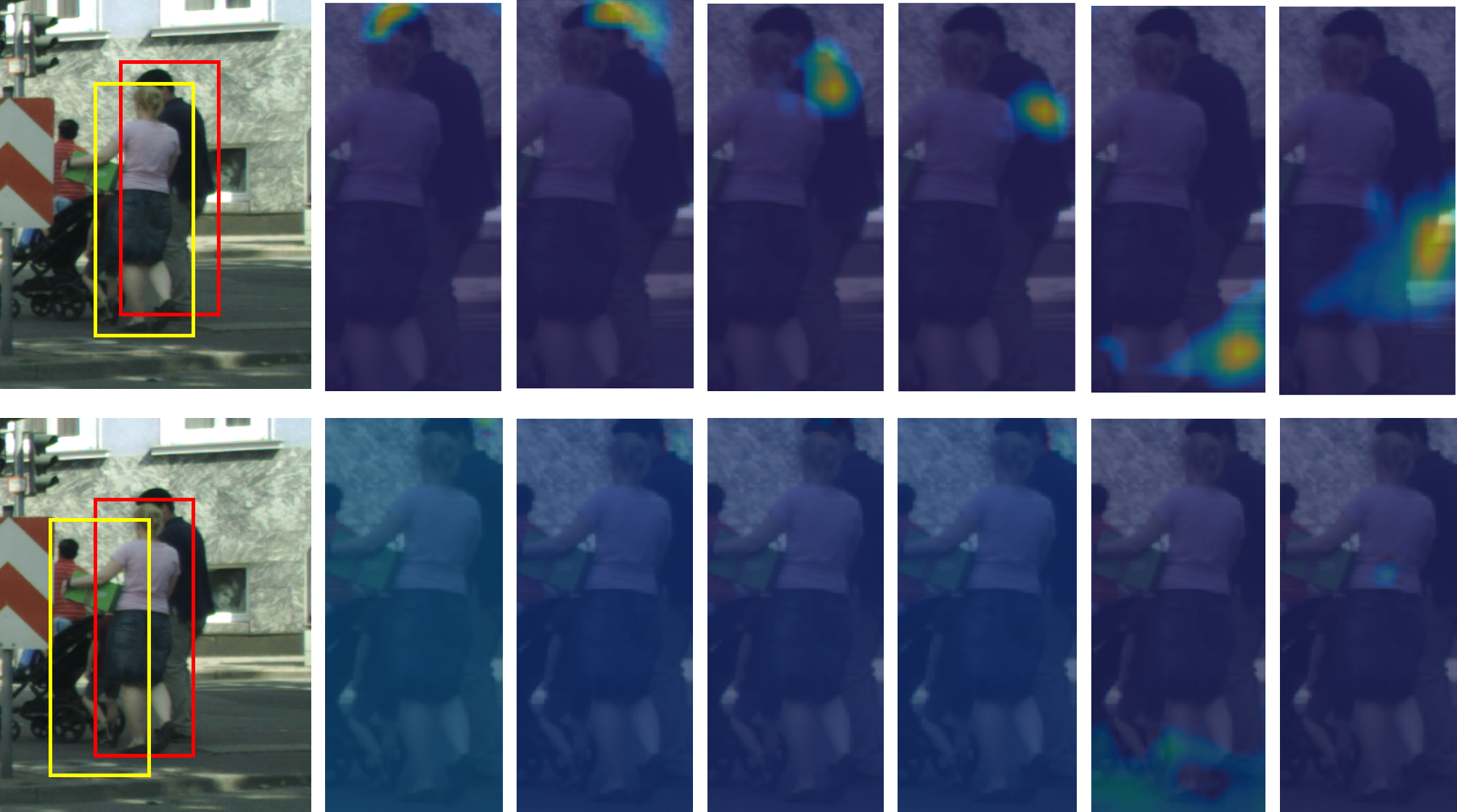}
		\caption{Visualization of correlation maps across channels between two proposals. Top row: Bob and Alice proposals, where high correlation values are only shown at visible parts of Bob. Bottom row: Alice and background proposals, where the correlation map only shows weak response.
		}
		\label{fig:vis_correlation}
	\end{figure}

	As shown in the top row of Fig.~\ref{fig:vis_correlation}, we find that for Bob high correlation values are only shown at his visible parts. This correlation map thus can be used to indicate visible parts of Bob.
	On the other hand, for fully visible Alice, we also observe the same channel features of her proposal and some arbitrary background proposal. We compute the correlation of two feature maps in the same way as above, but do not find high response regions, as shown in the bottom row of Fig.~\ref{fig:vis_correlation}. Although the background proposal also contains some body parts, they are not aligned with the target pedestrian and thus the two feature maps are not correlated. This observation indicates that high correlations are only present for aligned body parts between two proposals.
	
	Therefore, we conclude that the correlation maps between visible and occluded pedestrians can be used to identify occlusion patterns, without requiring additional information, \eg visible boxes or body part detections. 
	
	\subsubsection{Occluded pedestrian proposal selection}
	After obtaining the above correlation maps, we can then use them to locate occluded regions, which usually show low correlation values.
	Assume we have $X\times Y$ grids on the correlation map, for a given grid located at $(x_0,y_0)$ we take it as occluded region once its response value is below the average number of current correlation map:
	\begin{equation}
		c(x_0,y_0)<\frac{1}{X\times Y}\sum_{y=1}^{Y}\sum_{x=1}^{X}c(x,y),
	\end{equation}
	where $X, Y$ indicate the numbers of grids along horizontal and vertical, respectively. This dynamic criteria is more adaptive to different proposals with various feature values.
	
	Finally, the proposal is considered as occluded once the number of occluded grids exceeds a given threshold $\alpha$.


	\subsection{Adversarial Feature Completion}
	In the previous subsection, we have selected out occluded proposals. For each occluded proposal, occluded regions can be located by binarizing the correlation map with a given threshold $\beta$.
	
\begin{figure}
		\centering
		\includegraphics[width=0.5\textwidth]{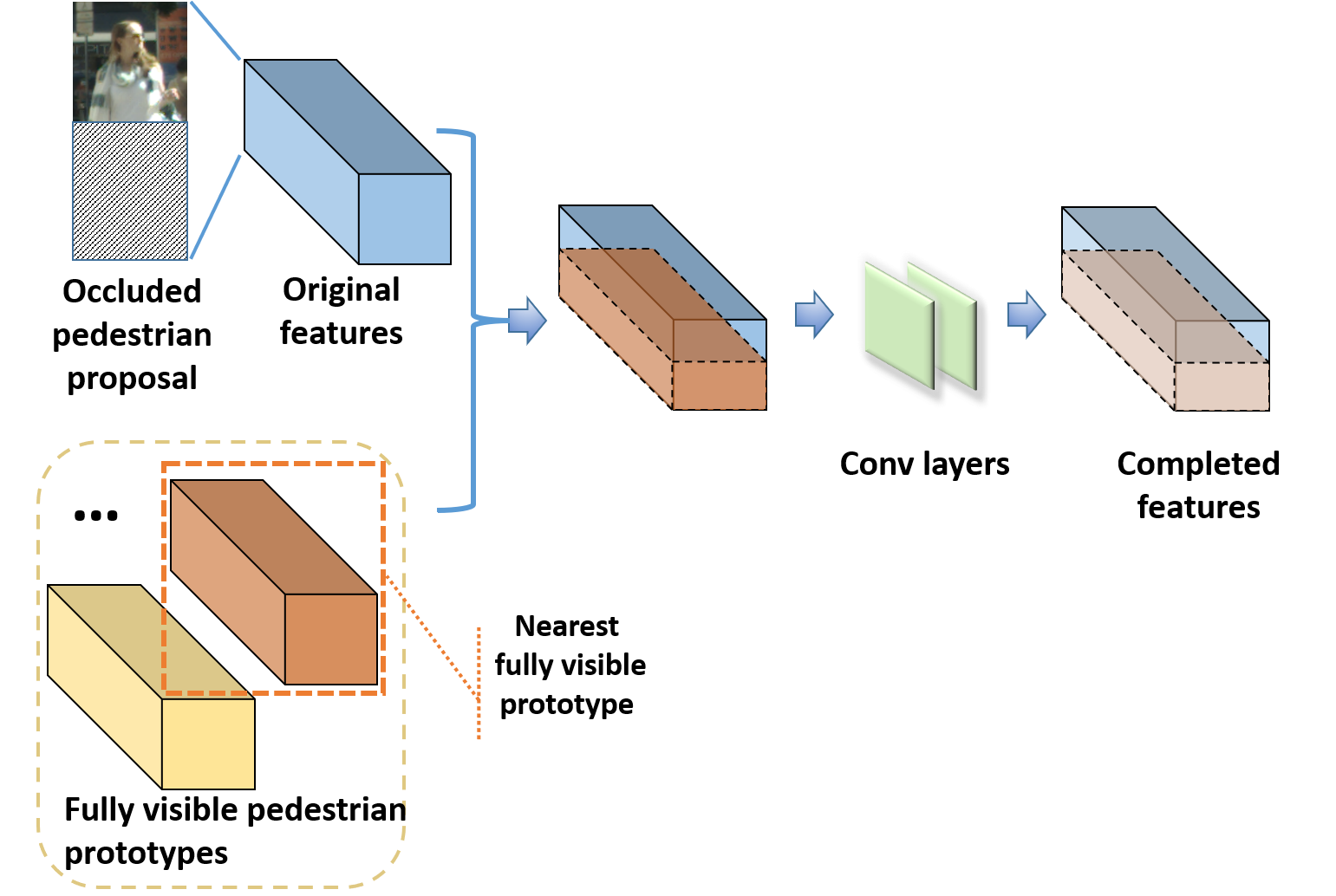}
		\caption{Illustration of the generator in our adversarial feature completion module.}
		\label{fig:generator}
	\end{figure}

	Since occluded regions contain occluding objects or pedestrians, those features are considered as noises and should be replaced by corresponding true body part features.
	An intuitive and simple way of feature completion is to borrow features from the nearest prototype at the above located occluded regions and fill the corresponding occluded regions at the occluded proposal, as illustrated in Fig. \ref{fig:generator}. In this way, the unseen regions at the occluded proposals are somehow completed.

	However, there would exist a big gap between the completed features and the real fully visible features due to the hard copy paste operation. This gap is harmful for the following classifier.
	Thus, we propose an adversarial learning module to close the above gap, making it easier for one coherent classifier to distinguish pedestrians of different occlusion patterns from background clutters.

	\begin{figure}
		\centering
		\includegraphics[width=0.5\textwidth]{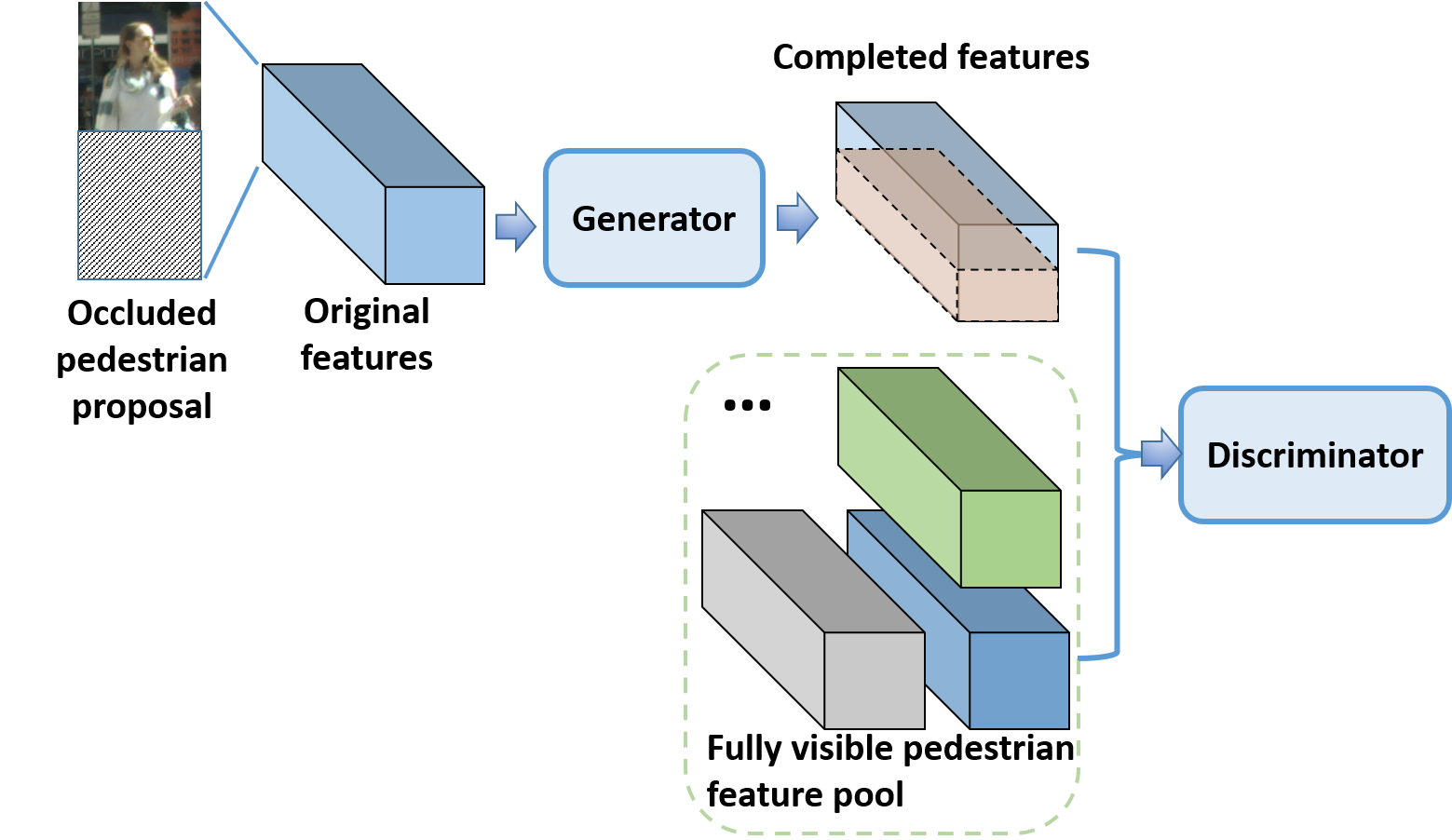}
		\caption{Overview of our adversarial feature completion module. The architecture of the generator is illustrated in Fig. \ref{fig:generator}. Please note the discriminator is discarded at test time.}
		\label{fig:GAN}
	\end{figure}
	
	\begin{algorithm}
		\caption{Adversarial Learning for Feature Completion \protect}
		\textbf{Input:} Occluded proposal feature pool $F_{occ}$ and fully visible proposal feature pool $F_{vis}$; learning rate $\gamma$.\\		
		\textbf{Initialize:} Model parameters $\Theta_G$ and $\Theta_D$ for the generator $G$ and discriminator $D$, respectively; number of total training iterations $T$; number of steps to apply to the discriminator $K$.
		\begin{algorithmic}[1]	
			\For{$t=1,2,...,T$} 
			\For{$k=1,2,...,K$} 
			\State Sample $m$ samples from $F_{occ}$;
			\State Sample $m$ samples from $F_{vis}$; 
			\State Update $D$ by ascending its stochastic gradient:
			\begin{align*} 
				\Theta_{D}^{k} &=\Theta_{D}^{k-1}+ \\	 &\gamma\bigtriangledown_{{\Theta}_{D}^{k-1}}\frac{1}{m}\sum_{i=1}^{m}[logD(F_{vis}^i)+log(1-D(G(F_{occ}^i)))];
			\end{align*}
			\EndFor
			\State $\Theta_{D}^{t}=\Theta_{D}^{K}$;
			\State Sample $m$ samples from $F_{occ}$;
			\State Update $G$ by descending its stochastic gradient:
			\[{\Theta}_{G}^{t}=\Theta_{G}^{t-1}-\gamma\bigtriangledown_{{\Theta}_{G}^{t-1}}\frac{1}{m}\sum_{i=1}^{m}log(1-D(G(F_{occ}^i)));\]
			\EndFor 
			\State \Return{$\Theta_G$ and $\Theta_D$}.
		\end{algorithmic}	
		\label{alg:adversarial}	
	\end{algorithm}

	The network structure of our adversarial learning module is shown in Fig. \ref{fig:GAN}. It is composed of a generator illustrated in Fig. \ref{fig:generator} and a discriminator with two fully connected layers.
	The generator and discriminator are optimized alternatively, following the strategy in \cite{GAN}. Please note the discriminator is only applied at training time while discarded at test time.
	The detailed training flow is described in Alg. \ref{alg:adversarial}.
	At each training iteration, we first optimize the discriminator with $K$ steps. At each step, $m$ samples from the occluded proposal feature pool $F_{occ}$ and $m$ samples from the fully visible proposal feature pool $F_{vis}$ are sampled to generate a minibatch, respectively; and then the discriminator is updated by ascending its stochastic gradient.
	After $K$ steps, the discriminator is frozen; we then sample another $m$ samples from the occluded proposal feature pool $F_{occ}$, and update the generator by descending its stochastic gradient. The above procedure is repeated until convergence.

	We notice that the gap between the completed occluded features and real fully visible pedestrian features is too big, caused by two factors: occlusions and different pedestrian identities. Such a big gap hinders the convergence of the completion module.
	To narrow down the gap, we propose to perform the above adversarial learning with the divide and conquer strategy, \ie to handle one factor after another.
	At the first step, we first mimic the occlusion patterns of real occluded samples, by applying the occlusion masks obtained from Sec. \ref{sec:Occlusion Pattern Modeling} on real fully visible pedestrians to generate synthetic occluded samples; then we complete features for each synthetic occluded sample as described above, and feed the pair of completed synthetic occluded features and its original real fully visible features to the discriminator. In this way, we handle the gap caused by occlusions for the same identity.
	At the second step, we use real occluded and real fully visible pedestrians to construct adversarial pairs, where larger difference induced by different identities is handled.
	Please note the only difference between two steps is how to construct  adversarial pairs, while the network structure is the same, as shown in Fig. \ref{fig:GAN}.
	In this progressive way, the generator is better optimized as the difficulties are distributed to two sequential steps.
	
	\section{Experiments}\label{sec:experiments}
	In this section, we first introduce the evaluation metrics we use, followed by a brief description to the datasets used in our experiments. We also provide some implementation details along with runtime analysis. After that, we report some experimental results, including analysis on the impact of our proposed feature completion method, ablation studies and comparisons to state-of-the-art methods.

	\subsection{Evaluation Protocol}
	We use the standard average-log miss rate (MR) in all experiments, which is computed in the FPPI range of $[10^{-2}, 10^{0}]$ \cite{Dollar2012PAMI}. 
	Since we handle occlusion in this paper, we show results across different occlusion levels, following the protocol in \cite{Dollar2012PAMI,CityPersons}:
	(1) Reasonable (\textbf{R}): visibility $\in [0.65, inf]$;
	(2) Heavy occlusion (\textbf{HO}): visibility $\in [0.20, 0.65]$;
	(3) Reasonable + Heavy occlusion (\textbf{R+HO}): visibility $\in [0.20, inf]$.
	
	The performance on the \textbf{R+HO} subset is used to measure the overall performance as it includes a wide range of occlusions.

	\begin{table}\small
		\renewcommand\arraystretch{1.5}
		\centering	
		\begin{tabular} { c | c |c | c | c |c}
			\hline 
			Dataset & \small{\#images} & \textbf{R} & \textbf{HO} & \textbf{R+HO} & Density\\ \hline\hline		
			\small{CityPersons (val.)} &500& 1579 & 733 & 2312 & 4.62\\
			Caltech (test) &4024 &1014 & 273 & 1287 & 0.32\\
			CrowdHuman (val.) &4370 & 61k & 34k &  95k &21.69\\
			\hline
		\end{tabular}
		\vspace{10pt}
		\caption{Comparison of the number of pedestrians on different evaluation subsets. The last column of density shows the number of persons per image on \textbf{R+HO}.}
		
		\label{tab:number_subsets}
	\end{table}
	\subsection{Datasets}
	\textbf{CityPersons.} We use the CityPersons dataset \cite{CityPersons} for most of our experiments. The CityPersons dataset was built upon the Cityscapes dataset \cite{Cordts2016Cityscapes}, which was recorded in multiple cities and countries across Europe and thus shows high diversity. More important, it includes a large number of crowded scenes as the data was recorded in the center of many big cities, \eg Cologne and Frankfurt. We use the original training, validation and test split, which are composed of 2975, 500 and 1525 images respectively.
	
	\noindent\textbf{Caltech.} The Caltech \cite{Dollar2012PAMI} dataset is one of the most popular ones for pedestrian detection. It consists of approximately 10 hours of $640\times 480$ 30Hz video taken from a vehicle driving through Los Angeles. Following \cite{CityPersons,pang2019mask-attention-pedestrian}, we sample with 10Hz from set00-set05 to get 42,782 training images. The 4,024 test images are sampled with 1Hz from set06-set10. 
	
	\noindent\textbf{CrowdHuman.} The CrowdHuman dataset \cite{shao2018crowdhuman} is collected by crawling images with the Google image
	search engine with approximately 150 query keywords. The whole dataset consists of around 25,000 images, from more than 40 different cities around the world, and covers various activities and numerous viewpoints. As its name suggests, the CrowdHuman dataset contains a large number of persons at crowds.

	\subsection{Implementation Details}
	
	When comparing with state-of-the-art methods, we name our proposed new detectors as \textbf{FeatComp} and \textbf{FeatComp++}, both of which are based on the FasterRCNN architecture, and with ResNet-50 and HRNet-32 as backbone respectively.

	On CityPersons, the baseline detector for fully visible prototype construction is trained with an initial learning rate of $2\times10^{-3}$ for 9 epochs and trained for another 5 epochs with a decreased learning rate of $2\times10^{-4}$. 
	For the first step of adversarial learning, we train with an initial learning rate of $2\times 10^{-3}$ for 3 epochs and with a decreased learning rate of $2\times 10^{-4}$ for another 3 epochs; for the second step, we train with an initial learning rate of $2\times 10^{-4}$ for 3 epochs and with a decreased learning rate of $2\times 10^{-5}$ for another 5 epochs.
	
	On Caltech, we use the CityPersons base model for fully visible prototype construction.
	For the first step of adversarial learning, we train with an initial learning rate of $10^{-6}$ for 4 epochs and with a decreased learning rate of $10^{-7}$ for another 2 epochs; for the second step, we train with an initial learning rate of $10^{-7}$ for 4 epochs and with a decreased learning rate of $10^{-8}$ for another 2 epochs.
	
	On CrowdHuman, the baseline detector for fully visible prototype construction is trained with an initial learning rate of $6.25\times 10^{-4}$ for 15 epochs and trained for another 4 epochs with a decreased learning rate of $6.25\times 10^{-5}$, following \cite{ProPred}. 
	For the first step of adversarial learning, we train with an initial learning rate of $6.25\times 10^{-4}$ for 5 epochs and with a decreased learning rate of $6.25\times 10^{-5}$ for another 5 epochs; for the second step, we train with an initial learning rate of $6.25\times 10^{-5}$ for 5 epochs and with a decreased learning rate of $6.25\times 10^{-6}$ for another 4 epochs.

	\myparagraph{Runtimes.} We report the runtimes of our detector by comparing to the baseline detector in Tab. \ref{tab:runtimes}. We run each detector on the entire CityPersons validation set and then take its average runtime per image. Our experiments are implemented on one Nvidia Geforce RTX 2080Ti GPU. As shown in the last column of Tab. \ref{tab:runtimes}, our proposed feature completion method only brings minimal additional runtimes compared to the baseline method, \ie $\sim$ 10 ms.

	\begin{table}
		\renewcommand\arraystretch{1.5}
		\centering	
		\begin{tabular} { c | c |c|c }
			\hline 
			Method & Scale & Runtime & $\Delta$ \\ \hline\hline		
			FasterRCNN &\multirow{2}{*}{1.0}& 166 ms & - \\
			FeatComp &&175 ms & +9 ms \\ \hline
			FasterRCNN &\multirow{2}{*}{1.3}& 280 ms & - \\
			FeatComp & &296 ms & +16 ms \\ 
			\hline
		\end{tabular}
		\vspace{10pt}
		\caption{Comparison of runtimes (average per image) between the baseline and our method on the CityPersons validation set. Our method only brings minimal additional runtimes compared to the baseline method, as shown in the last column.}	
		\label{tab:runtimes}
	\end{table}

	\subsection{Impact of Feature Completion}
	
	\begin{table*}
		\renewcommand\arraystretch{1.5}
		\centering	
		\begin{tabular} { c|c | c | c | c | c | c |c}
			\hline
			Detector & Feature & \multicolumn{2}{c|}{\textbf{HO}} & \multicolumn{2}{c|}{\textbf{R}} & \multicolumn{2}{c}{\textbf{R+HO}}   \\
			& completion&  MR & $\Delta$ MR & MR & $\Delta$ MR & MR & $\Delta$ MR \\
			\hline	\hline
			\multirow{2}{*}{FasterRCNN (VGGNet-16)}& $\times$ &49.52  & -& 12.90  & -  & 30.53 & -\\
			& $\surd$ &42.76& +6.76pp& 11.63 & +1.27pp   &26.16 & +4.37pp\\
			\hline		
			\multirow{2}{*}{FasterRCNN (ResNet-50)}& $\times$  & 45.85 & -&  12.84 & -  & 28.97& -\\
			& $\surd$&41.63& +4.22pp & 11.63 & +1.21pp   & 25.95& +3.02pp\\
			\hline		
			\multirow{2}{*}{FasterRCNN (DLA-34)} & $\times$  & 44.95 & - &  12.13 & - & 27.50& -\\
			& $\surd$&41.30& +3.65pp & 10.96 & +1.17pp   & 25.05& +2.45pp\\
			\hline	
			\multirow{2}{*}{FasterRCNN (HRNet-32)} & $\times$&39.92 & - & 10.99 & -  &24.30 & -\\
			& $\surd$ & 35.21& +4.71pp&9.79& +1.20pp   & 21.95 & +2.35pp\\
			\hline
			\multirow{2}{*}{D-DETR (ResNet-50)} & $\times$&39.01 & - & 11.61 & -  &25.31 & -\\
			& $\surd$ & 36.11& +2.90pp&10.62& +0.99pp   & 23.94 & +1.37pp\\
			\hline
		\end{tabular}
		\vspace{10pt}
		\caption{Applying our proposed feature completion approach on top of five different baseline detectors: four FasterRCNN detectors with different backbones, and one Transformer based detector D-DETR. Numbers are MR on the CityPersons validation set, lower is better.
			\textbf{R}: Reasonable; \textbf{HO}: Heavy occlusion; \textbf{R+HO}: Reasonable+Heavy occlusion. 
		}
		\label{tab:multi_baslines}
	\end{table*}
	
	We first validate the efficacy of our method by applying our proposed feature completion module on top of two kinds of baseline detectors: FasterRCNN with four different backbones: VGGNet-16 \cite{VGG}, ResNet-50 \cite{Resnet}, DLA-34 \cite{DLA} and HRNet-32 \cite{HRNet}; and Transformer based detector D-DETR \cite{D-DETR}. 
	We show the results on the CityPersons validation set in Tab.~\ref{tab:multi_baslines}, and we have the following observations:
	
	(1) Our proposed feature completion module brings significant improvements to their baselines on the heavy occlusion subset. Specifically, the improvements to five different baseline detectors on the heavy occlusion subset are 6.76pp, 4.22pp, 3.65pp, 4.71pp and 2.90pp respectively. These results indicate our feature completion module generates more classification-friendly features for occluded pedestrians, and thus the detection accuracy is improved.
	
\begin{figure}
		\centering
		\subfigure[]{\includegraphics[width=0.4\textwidth] {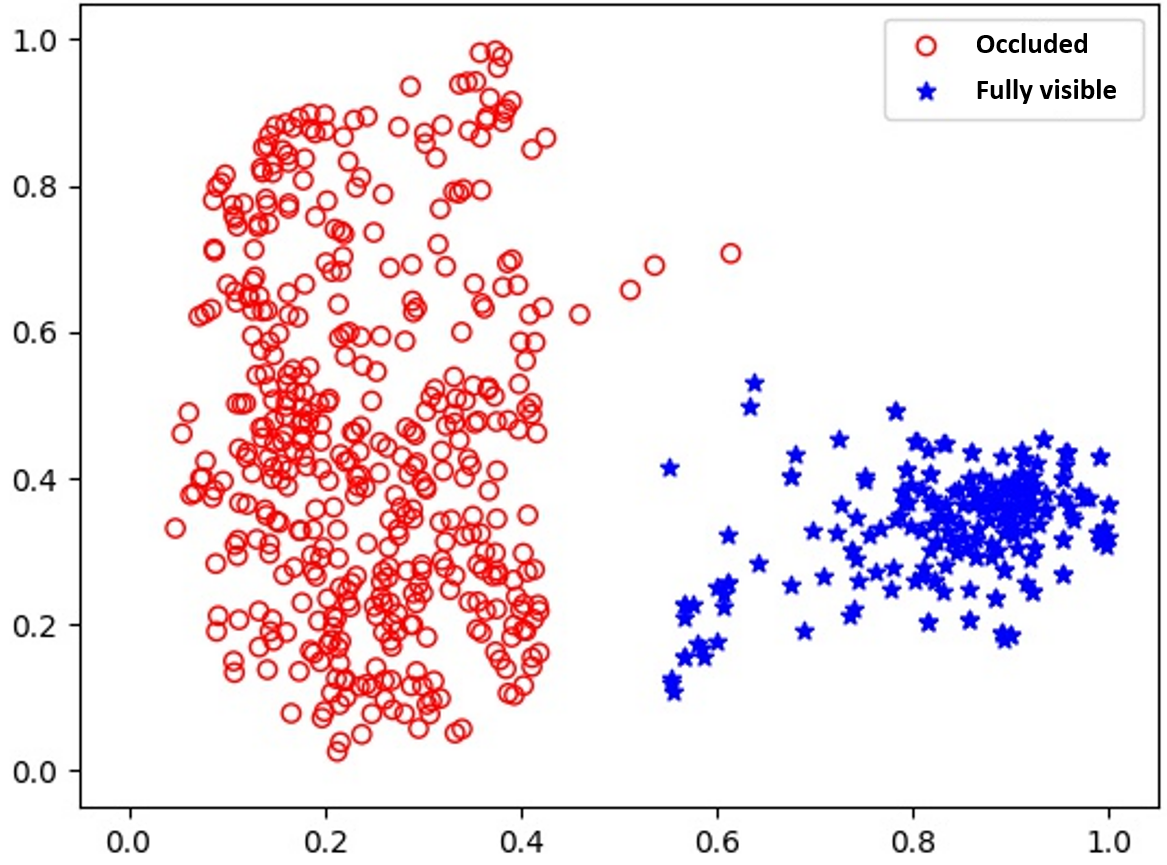}\label{fig:ftr_vis_withoutcomp}}
		\subfigure[]{\includegraphics[width=0.4\textwidth] {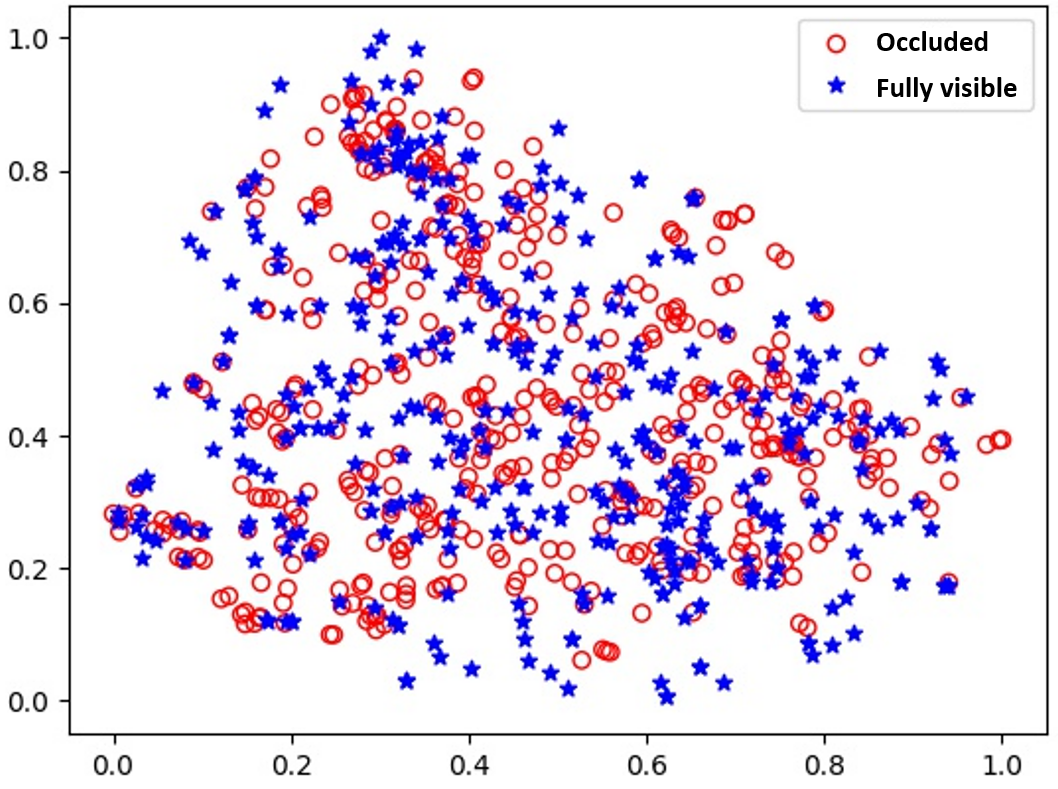}\label{fig:ftr_vis_withcomp}}
		\caption{Feature space visualization via t-SNE for all pedestrian samples on the CityPersons validation set. (a) In our baseline detector, the entire feature space is sparse as occluded pedestrian features stay far away from fully visible ones; (b) in our proposed detector FeatComp, the features of occluded and fully visible pedestrians are well aligned, generating a compact feature space.}
		\label{fig:feature space visualization}
	\end{figure}

	(2) We improve not only on the heavy occlusion subset, but also on the reasonable subset, where fully visible and partially occluded pedestrians are included. The relative improvements for five detectors are all around $10\%$.
	These notable gains demonstrate that our proposed method is able to handle different levels of occlusion by aligning the entire feature space of all pedestrians. We compare the feature spaces before and after our proposed feature completion module in Fig. \ref{fig:feature space visualization}, where we can see that for our baseline detector, the entire feature space is sparse as occluded pedestrian features stay far away from fully visible ones; in contrast, our proposed detector FeatComp well aligns the features of occluded and fully visible pedestrians, generating a compact feature space.
	The more compact feature space benefits classification between pedestrians and background clutters, thus improving the detection accuracy.
	
	(3) Our proposed method improves consistently over strong baselines.
	For example, FasterRCNN (HRNet-32) and D-DETR are stronger baselines, outperforming previous ones by more than 5pp on \textbf{HO}, but our proposed method still brings significant improvements, \ie 4.71pp and 2.90pp, respectively. 
	These consistent improvements indicate that our method is generalizable and can be applied on top of future detectors to achieve further improvements.

	\subsection{Ablation Studies}
	All the experiments in this subsection are implemented on the CityPersons validation set using FasterRCNN as detector architecture, ResNet-50 as backbone and the original image scale as input.

	\myparagraph{Number of fully visible pedestrian feature prototypes.}
	Since we employ fully visible pedestrian feature prototypes for feature completion, we study the impact of number of prototypes on the detection performance.
	As shown in Tab. \ref{tab:results_ablation_number_of_prototypes}, the detection performance varies slightly ($\sim$0.2pp on \textbf{R+HO}) when we use different numbers of prototypes. This is another evidence showing that pedestrian features are highly correlated such that it does not matter much when we use different prototypes for feature completion.
	Given the small gap among different numbers of prototypes, one may wonder whether it is possible to use some arbitrary features for completion, \ie getting rid of fully visible pedestrian feature prototypes. To answer this question, we further implement an additional experiment, where we use random features from the background clutters for completion, and find that it results in very bad results, even worse than the baseline detector. This observation further verifies that background features are not correlated with pedestrian features.
	Therefore, we conclude it is necessary to use fully visible pedestrian feature prototypes for feature completion, but the performance is stable with prototype construction methods.
	
	\begin{table}
		\renewcommand\arraystretch{1.5}
		\centering
		\begin{tabular} { c| c | c | c }
			\hline
			\# Prototypes & \textbf{HO} & \textbf{R}  & \textbf{R+HO}  \\
			\hline	\hline		
			3 & 39.70& 11.74  &25.79 \\ 
			4 &39.30&11.91  &25.75\\
			5 &\textbf{39.07}&11.61  &\textbf{25.63}\\
			6 &39.15&11.69  &25.71\\
			7 &39.55&\textbf{11.57}  &25.76\\
			\hline
			None &49.52& 13.19  &30.99 \\
			\hline
			Baseline &45.85& 12.84  &27.97 \\
			\hline
		\end{tabular}
		\vspace{10pt}
		\caption{Ablation study on number of fully visible pedestrian feature prototypes. Here we use ResNet-50 as the backbone.}
		\label{tab:results_ablation_number_of_prototypes}
	\end{table}

\begin{figure}
		\centering
		\includegraphics[width=0.4\textwidth]{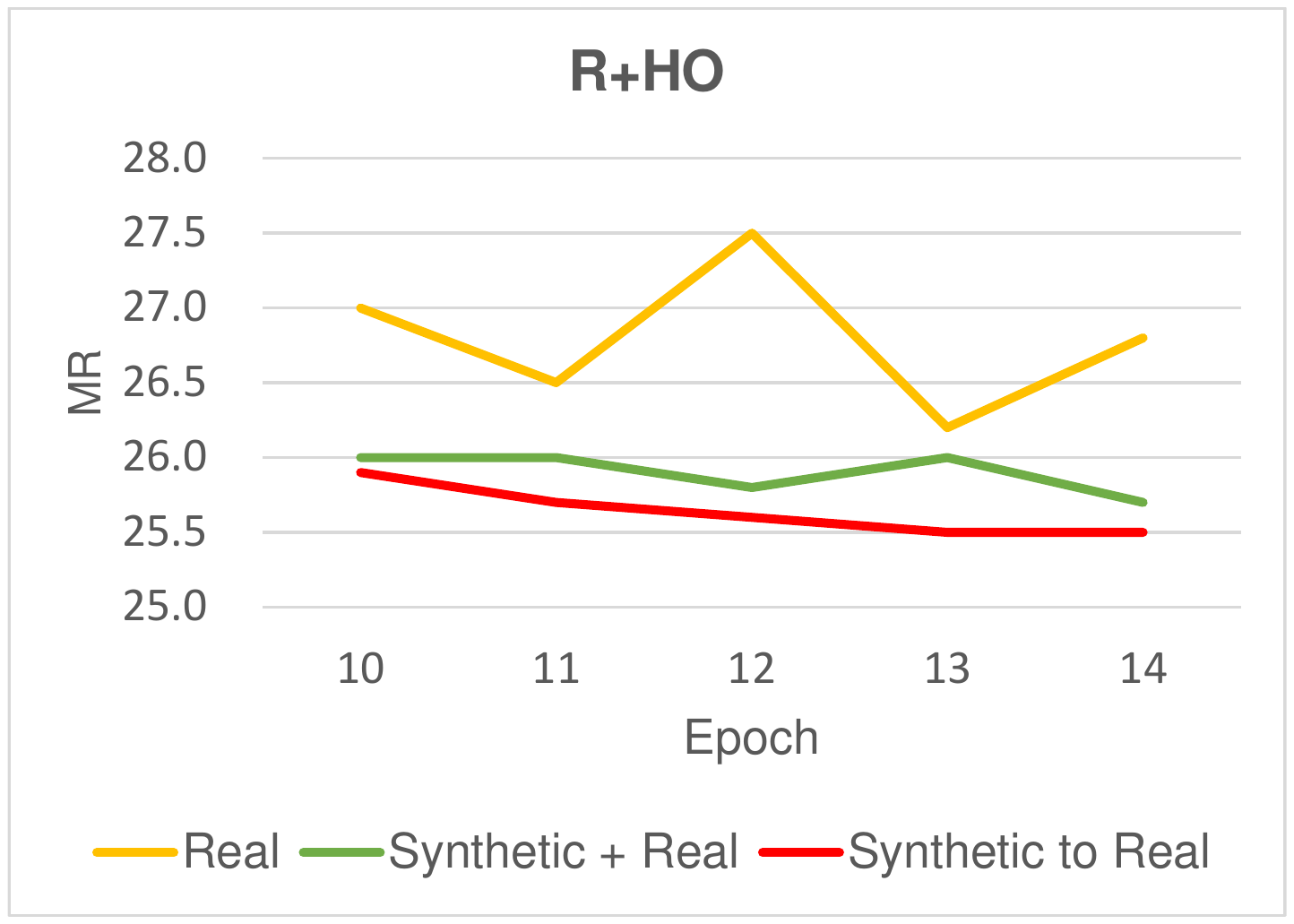}
		\caption{Comparison of different adversarial learning strategies. Our progressive strategy allows the model to converge better, providing better and more stable results.}
		\label{fig:ablation_progressive_adv}
	\end{figure}

	\myparagraph{Progressive adversarial learning.}
	In order to better optimize the generator for adversarial feature completion, we propose a progressive learning strategy, \ie, synthetic and real occluded samples are used at the first and second step respectively. Here we study the effect of this progressive learning strategy. In Fig. \ref{fig:ablation_progressive_adv}, we show the miss rates over different training stages. When we directly use real occluded samples for adversarial learning, the optimization procedure of the model seems unstable; when we combine synthetic with real occluded samples for adversarial learning, the performance improves, but the stability is still poor; in contrast, our progressive learning strategy allows the model to converge better, providing better and more stable results.
	
	\begin{table*}
		\centering
		\renewcommand\arraystretch{1.5}	
		\begin{tabular} { c || c|c || c | c | c || c |c | c }
			\hline
			Detector & Backbone & Extra & \multicolumn{3}{c||}{Validation} & \multicolumn{3}{c}{Test} \\
			
			& &&\textbf{HO} & \textbf{R} &  \textbf{R+HO} &  \textbf{HO}&\textbf{R}  & \textbf{ALL} \\
			
			\hline	\hline
			RepLoss \cite{RepulsionLoss}& ResNet-50 & $\times$  &54.78&11.57   & 39.83 &52.59&11.48   & 39.17\\
			Adaptive-NMS \cite{liu2019adaptive-nms-pedestrian}  &VGGNet-16	& $\times$  &54.01& 10.83   &40.47 &46.99 & 11.40   & 38.89\\	
			Adapted FasterRCNN \cite{CityPersons} &VGGNet-16 & $\times$
			&53.48&12.81  &36.81 &50.47&12.97   & 43.86  \\
			PBM \cite{PBM} &VGGNet-16& vbb  & 53.30& 11.10  & -& -& -& -\\
			RelationNet \cite{RelationNetworks}& VGGNet-16 &$\times$ &52.60 & 12.75  &35.63 &50.71&12.30&45.96 \\
			OR-CNN \cite{OcclusionAware} &VGGNet-16 & vbb &51.90&10.94   &39.41  &51.43&11.32   & 40.19 \\
			FasterRCNN+ATT \cite{AttentionNet}&VGGNet-16 &parts & 48.62 &13.25  & 33.01 & - & - & -\\
			APD \cite{APD}& DLA-34 &$\times$ &46.60 &8.80 & - &35.45 &8.27 &35.65 \\
			Bi-box \cite{Bi-box}& VGGNet-16 &$\times$ &44.20 &11.20   &- &- &- &- \\		
			FRCN+A+DT \cite{zhou2019feature-transformation-pedestrian}&VGGNet-16 & $\times$ &44.30 &11.10   & - &- &- &-\\	
			OAF-Net \cite{OAF-Net}& HRNet-32 & vbb &43.10 &9.40   &- &- &- &- \\
			PRNet \cite{PRNet}&ResNet-50 & vbb &42.00 &10.80   &25.60 &- &- &- \\
			HGPD \cite{HGPD}& VGGNet-16 &$\times$ &40.90 & 10.90  &- &38.65&10.17&38.24 \\
			CSP \cite{liu2019anchor-free-pedestrian} &ResNet-50 & $\times$  & 40.56&11.00 &  26.13 &- &- &-\\
			
			MGAN \cite{pang2019mask-attention-pedestrian} &VGGNet-16 & vbb  &39.40&{10.50}   & - &40.97&9.29   & 38.86\\
			VLPD \cite{VLPD_2023_CVPR}& ResNet-50 &language &34.90 &9.40   &21.70 &- &- &- \\
			F2DNet \cite{F2DNet}&HRNet-32 & $\times$ &32.60 &8.70   &- &- &- &- \\
			LSFM \cite{LSFM_2023_CVPR}&HRNet-32 & $\times$ &31.90 &8.50 &-  &- &- &-  \\
			\hline
			FeatComp&ResNet-50 & $\times$ &{38.09}  & {10.08}  &  {23.52} &{37.50}&{9.15}  &{37.85}\\
			FeatComp++&HRNet-32 & $\times$ &\textbf{31.78}  & \textbf{7.61}  &  \textbf{19.54} &\textbf{32.96}&\textbf{7.37}  &\textbf{34.86}\\
			\hline
		\end{tabular}
		\vspace{10pt}
		\caption{Comparison to state-of-the-art methods on the CityPersons dataset. Numbers indicate MR, lower is better. Some methods use different definitions of \textbf{HO} in their papers, and we re-evaluate under a consistent setting for a fair comparison.
			Test \textbf{ALL}:visibility $\in [0.20,inf]$, height $\in [20,inf]$. Results generated by using external training data are not included.}
		\label{tab:results_citypersons}
	\end{table*}

	\subsection{Comparisons to State-of-the-art Methods}
	\myparagraph{CityPersons.}
	We compare our methods to other state-of-the-art detectors in Tab.~\ref{tab:results_citypersons}.
	We find our detector FeatComp using ResNet-50 outperforms previous top method MGAN \cite{pang2019mask-attention-pedestrian}, which relies on visible boxes as an extra cue.
	We note quite some recent methods, \eg OAF-Net \cite{OAF-Net}, F2DNet \cite{F2DNet}, LSFM \cite{LSFM_2023_CVPR}, use HRNet-32 to obtain better results. For fair comparison, we also compare with them using our detector FeatComp++. Please note we do not include those results generated by using external training data considering fairness, for example, LSFM \cite{LSFM_2023_CVPR} results on the test set, and the results from Pedestron \cite{Ped_Det_Elephant} and DIW loss \cite{DIW_2022_CSCS}.
	We observe that our detector FeatComp++ achieves better results than previous methods, in the way that we better balance the performance at different occlusion levels.
	In the end, our detector FeatComp++ establishes a new state of the art on both validation and test sets, even surpassing previous methods that use extra cues and annotations, \eg visible boxes (OR-CNN \cite{OcclusionAware}, MGAN \cite{pang2019mask-attention-pedestrian}, PBM \cite{PBM}), part detections (FasterRCNN+ATT \cite{AttentionNet}) and vision-language large models (VLPD \cite{VLPD_2023_CVPR}). These results indicate that our occlusion modeling method is effective.

\begin{figure*}[t]
		\centering
		\includegraphics[width=\textwidth]{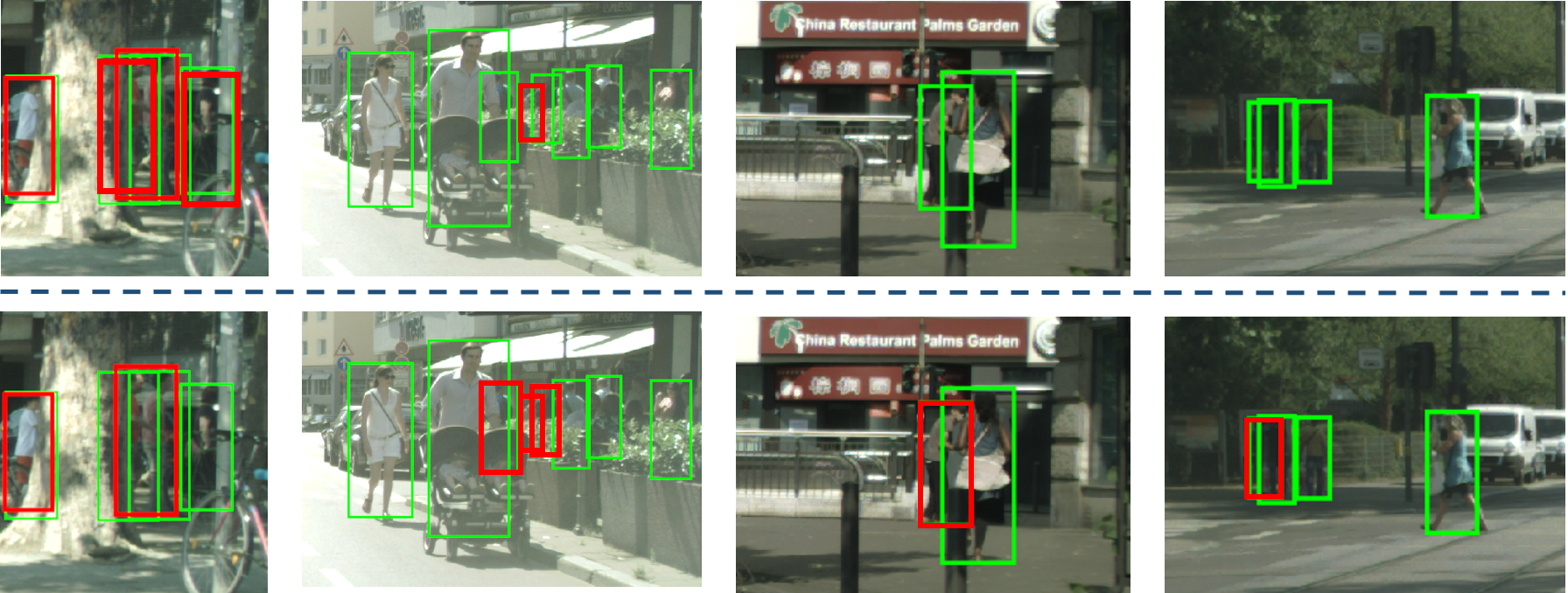}
		\caption{Qualitative results on the CityPersons validation set from our FeatComp detector (upper row) and the baseline FasterRCNN (ResNet-50) detector (lower row). Our detector using feature completion module achieves a higher recall for occluded pedestrians. We show \textcolor{green}{true positive} detections in green and \textcolor{red}{missing recall} in red; the results are from FPPI=0.1. 
		}
		\label{fig:vis_dt2}
	\end{figure*}

	We further show some qualitative results at crowded scenes in Fig.~\ref{fig:vis_dt2}. Compared to the baseline detector, our method achieves a higher recall for occluded pedestrians, demonstrating that our method handles different kinds of occlusion effectively.

	\myparagraph{Caltech}. We also report experimental results on Caltech, which is a standard benchmark for pedestrian detection. We compare with other state-of-the-art methods in Tab.~\ref{tab:results_caltech}.
	We note that our FeatComp detector obtains comparable results to a recent detector OPL \cite{OPL_2023_CVPR} on \textbf{HO}, yet better results on the overall set of \textbf{R+HO}, indicating that our detector is better at handling samples of different occlusion levels.
	We even outperform those methods using extra cues like visible boxes (JL-TopS \cite{MultiLabel-occl}, PCN \cite{PCN}), body part detections (FasterRCNN+ATT \cite{AttentionNet}), temporal information (TLL-TFA \cite{TopologicalLine}), and vision-language large models (VLPD \cite{VLPD_2023_CVPR}).
	Furthermore, our detector FeatComp++ establishes a new state of the art on Caltech, surpassing the previous top method by around 6pp on \textbf{HO}.

	\begin{table}
		\centering
		\renewcommand\arraystretch{1.5}	
		\begin{tabular} { c ||c|| c | c | c }
			\hline
			Detector  & Extra & \textbf{HO} & \textbf{R}  & \textbf{R+HO}  \\
			\hline	\hline	        
			RPN+BF\cite{rpn+bf16eccv} & $\times$ &57.74& 7.28  &20.00 \\
			DeepParts \cite{DeepParts} & $\times$&56.27 & 12.90    &24.38 \\
			MS-CNN\cite{cai16mscnn} & $\times$ &48.66& 8.08  &18.59 \\
			JL-TopS \cite{MultiLabel-occl}& vbb& 43.67 & 7.59   & 16.73 \\ 
			PCN \cite{PCN}& vbb& 43.53 & 8.47   & 17.45 \\
			AR-Ped \cite{brazil2019autogressive-pedestrian}& $\times$  &43.46 &{5.13}  & 15.14 \\
			SDS-RCNN \cite{SDS-RCNN}& $\times$& 42.56 & 6.44   & 16.39 \\
			TLL-TFA \cite{TopologicalLine}& temporal& 40.91 & 12.39   & 20.61 \\ 
			GDFL \cite{Graininess}& $\times$ & 38.76& 6.32   & 14.88 \\
			F2DNet \cite{F2DNet}& $\times$ &38.70 &2.20   &-\\
			FRCN+A+DT \cite{zhou2019feature-transformation-pedestrian}& $\times$ &37.90 &8.00   & - \\
			VLPD \cite{VLPD_2023_CVPR} & language & 37.70&2.30&-\\
			LSFM \cite{LSFM_2023_CVPR}& $\times$ &35.80 &3.10   &-\\
			\small{FasterRCNN+ATT} \cite{AttentionNet}  & parts &  37.18 & 8.11   &  15.89 \\					
			HGPD \cite{HGPD}& $\times$ &36.33 & 4.83  &- \\
			
			TCE \cite{TCE}& temporal & 30.90& 6.70   & 12.40 \\
			OPL \cite{OPL_2023_CVPR}& $\times$ & 30.10& 5.20   & 11.70 \\
			\hline
			\small{FeatComp} & $\times$ &  {30.74}&{4.03} &  {10.53}\\
			\small{FeatComp++} & $\times$ &  \textbf{24.55}&\textbf{2.75} &  \textbf{7.91}\\ 
			\hline
			
		\end{tabular}
		\vspace{10pt}
		\caption{Comparison to state-of-the-art detectors on the Caltech test set. Numbers indicate MR evaluated on the new annotations from \cite{shanshan_cvpr16}; the second column indicates whether some extra cues have been used.}
		
		\label{tab:results_caltech}
	\end{table}
	
	\myparagraph{CrowdHuman}. We further show experimental results on the CrowdHuman validation set in Tab.~\ref{tab:results_crowdhuman}.
	CrowdHuman is considered as a more challenging dataset as the density is pretty high as shown in Tab. \ref{tab:number_subsets}.
	We compare with state-of-the-art methods, and find that our FeatComp detector achieves comparable results with previous top methods, \eg AEVB \cite{AEVB} and AutoPedestrian \cite{AutoPedestrian}. Furthermore, our FeatComp++ detector establishes a new state of the art, surpassing previous methods by more than 3pp.
	Please note that our methods not only outperform those methods rely on extra cues of visible boxes, \eg OAF-Net \cite{OAF-Net}, PEDR \cite{PEDR} and PBM \cite{PBM}, but also achieves better results than those recently proposed Transformer based methods, \eg DETR \cite{DETR}, D-DETR \cite{D-DETR} and D-DETR+ProPred \cite{ProPred}.

	\begin{table}
		\renewcommand\arraystretch{1.5}
		\centering
		\begin{tabular} { c||c|c  }
			\hline
			Detector & Extra cue&  MR    \\
			\hline	\hline
			DETR \cite{DETR}& $\times$& 73.20 \\
			RetinaNet \cite{shao2018crowdhuman} & $\times$& 63.33 \\
			FCOS \cite{FCOS} & $\times$& 54.00 \\
			ATSS \cite{ATSS} & $\times$& 51.10 \\
			FPN \cite{shao2018crowdhuman} & $\times$&  50.42 \\
			Adaptive-NMS \cite{liu2019adaptive-nms-pedestrian} & $\times$& 49.73 \\
			POTO \cite{POTO}&$\times$ & 47.80 \\
			OAF-Net \cite{OAF-Net}&vbb& 45.00 \\
			OPL \cite{OPL_2023_CVPR}& $\times$ & 44.90\\
			S-RCNN \cite{S-RCNN}&$\times$ & 44.70 \\
			DMSFLN \cite{DMSFLN}&$\times$ & 43.60 \\
			D-DETR \cite{D-DETR}&$\times$& 43.70 \\
			PEDR \cite{PEDR}&vbb & 43.70 \\
			PBM \cite{PBM}&vbb& 43.35 \\
			PGD \cite{PGD}&$\times$ & 42.80 \\	
			D-DETR+ProPred \cite{ProPred}&$\times$ & 41.50 \\
			MIP \cite{MIP} &$\times$ & 41.40 \\
			AEVB \cite{AEVB}&$\times$ & 40.70 \\
			AutoPedestrian \cite{AutoPedestrian}&$\times$ & 40.60 \\
			\hline	
			FeatComp& $\times$ &  40.89 \\
			FeatComp++& $\times$ & \textbf{37.46}  \\	
			\hline
		\end{tabular}
		\vspace{10pt}
		\caption{Comparison to state-of-the-art detectors on the CrowdHuman validation set. Bold indicates the best results.}
		\label{tab:results_crowdhuman}
	\end{table}

	\section{Conclusion}
	In this paper, we focus on the problem of occluded pedestrian detection. First, we point out occlusion increases the intra-class variance of pedestrians, hindering the model from finding the accurate classification boundary between pedestrians and background clutters. From the perspective of reducing the notable intra-class variance, we propose an intuitive method, which is to perform feature completion for occluded regions so as to align the features of pedestrians across different occlusion patterns.
	
	In order to perform effective feature completion, we have two key problems to solve. The first one is how to locate occluded regions; and the second one is how to close the gap between artificially completed features and real fully visible ones.
	
	To solve the first problem, we analyze some occluded pedestrian proposals and find that, channel features of different proposals show high correlation values at visible parts yet low correlation values at invisible parts. Motivated by these findings, we propose to use feature correlations to model occlusion patterns. The biggest advantage of such a modeling method is that it does not rely on extra cues or annotations such as visible boxes or part detections.
	
	To solve the second problem, we propose an adversarial learning method, which completes occluded features with a generator such that they can hardly be distinguished by the discriminator from real fully visible features.
	
	We report experimental results on the CityPersons, Caltech and CrowdHuman datasets. On CityPersons, we show significant improvements over five different baseline detectors with different backbones and detection architectures, especially on the heavy occlusion subset. Furthermore, we show that our proposed method FeatComp++ establishes a new state of the art on all the above three datasets.
	
	\section*{Data Availability Statement}
	The datasets used in this work are publicly available
	at the following links:
	
	\begin{enumerate}
		\item CityPersons. \url{https://github.com/cvgroup-njust/CityPersons}
		\item Caltech. \url{https://data.caltech.edu/records/f6rph-90m20}
		\item CrowdHuman. \url{https://www.crowdhuman.org/}
	\end{enumerate}

	\bibliographystyle{spbasic}      
	\footnotesize
	\bibliography{egbib}
	
\end{document}